\documentclass[conference]{IEEEtran}
\usepackage{xcolor}

\ifCLASSINFOpdf
\usepackage[pdftex]{graphicx}
\else
\fi
\usepackage{amsmath}
\usepackage{amssymb}

\usepackage{url}

\begin{document}
%
\title{Deep Reinforcement Fuzzing}


\author{\IEEEauthorblockN{Konstantin B\"ottinger$^{1}$, Patrice Godefroid$^2$, and Rishabh Singh$^2$}
\IEEEauthorblockA{$^1$Fraunhofer AISEC, 85748 Garching, Germany\\konstantin.boettinger@aisec.fraunhofer.de\\
$^2$Microsoft Research, 98052 Redmond, USA\\\{pg,risin\}@microsoft.com}}

\maketitle

\begin{abstract}

Fuzzing is the process of finding security vulnerabilities in
input-processing code by repeatedly testing the code with modified
inputs. In this paper, we formalize fuzzing as a reinforcement
learning problem using the concept of Markov decision processes. This
in turn allows us to apply state-of-the-art deep $Q$-learning
algorithms that optimize rewards, which we define from runtime
properties of the program under test. By observing the rewards caused
by mutating with a specific set of actions performed on an initial
program input, the fuzzing agent learns a policy that can next
generate new higher-reward inputs. We have implemented this new
approach, and preliminary empirical evidence shows that reinforcement
fuzzing can outperform baseline random fuzzing.

\end{abstract}


%
\IEEEpeerreviewmaketitle

\section{Introduction}

{\em Fuzzing} is the process of finding security vulnerabilities in
input-processing code by repeatedly testing the code with modified, or
{\em fuzzed}, inputs. Fuzzing is an effective way to find security
vulnerabilities in software~\cite{sutton2007fuzzing}, and is becoming
standard in the commercial software development process~\cite{SDL}.

Existing fuzzing tools differ by how they fuzz program inputs, but
none can explore exhaustively the entire input space for realistic
programs in practice. Therefore, they typically use {\em fuzzing
heuristics} to prioritize what (parts of) inputs to fuzz next. Such
heuristics may be purely random, or they may attempt to optimize for a
specific goal, such as maximizing code coverage.

In this paper, we investigate how to formalize fuzzing as a reinforcement
learning problem. Intuitively, choosing the next fuzzing action given
an input to mutate can be viewed as choosing a next move in a game
like Chess or Go: while an optimal strategy might exist, it is unknown
to us and we are bound to play the game (many times) in the search for
it. By reducing fuzzing to reinforcement learning, we can then try to
apply the same neural-network-based learning techniques that have
beaten world-champion human experts in Backgammon
\cite{tesauro1992practical,tesauro1995td}, Atari games
\cite{mnih2015human}, and the game of Go \cite{silver2016mastering}.

Specifically, fuzzing can be modeled as learning process with a
feedback loop. Initially, the fuzzer generates new inputs, and then
runs the target program with each of them. For each program execution,
the fuzzer extracts runtime information (gathered for example by
binary instrumentation) for evaluating the quality (with respect to
the defined search heuristic) of the current input. For instance, this
quality can be measured as the number of (unique or not) instructions
executed, or the overall runtime of the execution. By taking this
quality feedback into account, a feedback-driven fuzzer can learn from
past experiments, and then generate other new inputs hopefully of
better quality. This process repeats until a specific goal is reached,
or bugs are found in the program.  Similarly, the reinforcement
learning setting defines an agent that interacts with a system. Each
performed action causes a state transition of the system. Upon each
performed action the agent observes the next state and receives a
reward. The goal of the agent is to maximize the total reward over
time.

\begin{figure}
	\centering
	\includegraphics[scale=.7]{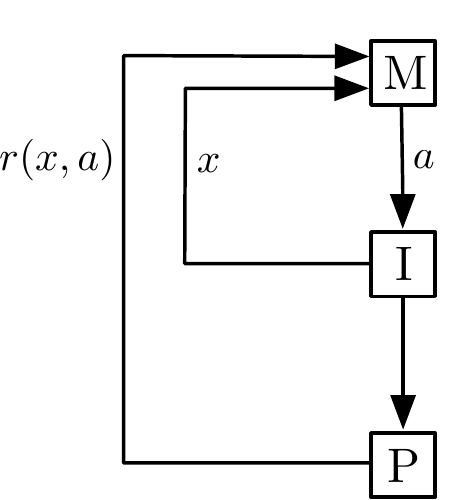}
	\caption{Modeling Fuzzing as a Markov decision process.}
	\label{fig:rf_architecture}
\end{figure}

Our mathematical model of fuzzing is captured in Figure~\ref{fig:rf_architecture}. An input mutator engine M generates a new input I by performing a fuzzing action $a$, and subsequently observes a new state $x$ directly derived from I as well as a reward $r(x,a)$ that is measured by executing the target program P with input I. We reduce input fuzzing to a reinforcement learning problem by formalizing it using Markov decision processes~\cite{sutton1998reinforcement}.
Our formalization allows us to apply state-of-the-art machine learning methods. In particular, we experiment with deep $Q$-learning.

In summary, we make the following contributions:
\begin{itemize}
	\item We formalize fuzzing as a reinforcement learning problem using the concept of Markov decision processes.
	\item We introduce a fuzzing algorithm based on deep $Q$-learning that learns to choose highly-rewarded fuzzing actions for any given propgram input.
	\item We implement and evaluate a prototype of our approach.
        \item We present empirical evidence that reinforcement fuzzing can outperform baseline random fuzzing.
\end{itemize}


\section{Related Work}
\label{sec:rf_related_work}
Our work is influenced by three main streams of research: fuzzing, grammar reconstruction, and deep $Q$-learning.

\subsection{Fuzzing}

There are three main types of fuzzing techniques in use today: (1)
{\em blackbox random}
fuzzing~\cite{sutton2007fuzzing,takanen2008fuzzing}, (2) {\em whitebox
constraint-based} fuzzing~\cite{godefroid_automated_2008}, and (3)
{\em grammar-based}
fuzzing~\cite{purdom1972sentence,sutton2007fuzzing}, which can be
viewed as a variant of model-based
testing~\cite{utting2006tmb}. Blackbox and whitebox fuzzing are fully
automatic, and have historically proved to be very effective at
finding security vulnerabilities in binary-format file parsers. In
contrast, grammar-based fuzzing is not fully automatic: it requires an
input grammar specifying the input format of the application under
test. This grammar is typically written by hand, and this process is
laborious, time consuming, and error-prone.  Nevertheless,
grammar-based fuzzing is the most effective fuzzing technique known
today for fuzzing applications with complex structured input formats,
like web-browsers which must take as (untrusted) inputs web-pages
including complex HTML documents and JavaScript code.

State-of-art fuzzing tools like SAGE~\cite{godefroid_automated_2008}
or AFL~\cite{afl} use coverage-based heuristics to guide their search
for bugs towards less-covered code parts. But they do not use machine
learning techniques as done in this paper.


Combining statistical neural-network-based machine learning with fuzzing is a novel approach and, to the best of our knowledge, there is just one prior paper on this topic: Godefroid et al.~\cite{learnfuzz-machine-learning-input-fuzzing} use character-based language models to learn a generative model of fuzzing inputs, but they do not use reinforcement learning.

\subsection{Grammar Reconstruction}
Research on reconstructing grammars from sample inputs for testing purposes started in the early 1970's \cite{purdom1972sentence,hanford1970automatic}. More recently, Bastani et al. \cite{Bastani:2017:SPI:3062341.3062349} proposed an algorithm for automatic synthesis of a context-free grammar given a set of seed inputs and a black-box target. Cui et al. \cite{Cui:2008:TAR:1455770.1455820} automatically detect record sequences and types in the input by identification of chunks based on taint tracking input data in respective subroutine calls. Similarly, the authors of \cite{Clause:2009:PAI:1572272.1572301} apply dynamic tainting to identify failure-relevant inputs. Another recently proposed approach~\cite{Hoschele:2017:MIG:3098344.3098355} mines input grammars from valid inputs based on feedback from dynamic instrumentation of the target by tracking input characters.

\subsection{Deep $Q$-Learning}
Reinforcement learning \cite{szepesvari2010algorithms} emerged from trial and error learning and optimal control for dynamic programming \cite{sutton1998reinforcement}. Especially the $Q$-learning approach introduced by Watkins \cite{wattkins1989learning,watkins1992q} was recently combined with deep neural networks \cite{tesauro1992practical,tesauro1995td,mnih2015human,silver2016mastering} to efficiently learn policies over large state spaces and has achieved impressive results in complex tasks. 


\section{Reinforcement Learning}
\label{sec:rf_background}

In this section we give the necessary background on reinforcement learning. We first introduce the concept of Markov decision processes \cite{sutton1998reinforcement}, which provides the basis to formalize fuzzing as a reinforcement learning problem. Then we discuss the $Q$-learning approach to such problems and motivate the application of deep $Q$-networks.

Reinforcement learning is the process of adapting an agent's behavior during interaction with a system such that it learns to maximize the received rewards based on performed actions and system state transitions. The agent performs actions on a system it tries to control. For each action, the system undergoes a state transition. In turn, the agent observes the new state and receives a reward. The aim of the agent is to maximize its cumulative reward received during the overall time of system interaction. The following formal notation relates to the presentation given in \cite{szepesvari2010algorithms}.


The interaction of the agent with the system can be seen as a stochastic process. In particular, a Markov decision process $\mathcal{M}$ is defined as $\mathcal{M}=(X,A,P_0)$, where $X$ denotes a set of states, $A$ a set of actions, and $P_0$ the transition probability kernel. For each state-action pair $(x,a) \in X\times A$ and each $U \subset X\times \mathbb{R}$  the kernel $P_0$ gives the probability	$P_0(U | x,a)$ such that performing action $a$ in state $x$ causes the system to transition into some state of $X$ that yields some real-valued reward $U$. $P_0$ directly provides the state transition probability kernel $P$ for single transitions $(x,a,y) \in X \times A \times X$
\begin{align}
	P(x,a,y)=P_0(\{y\}\times \mathbb{R} | x,a).
\end{align}
This naturally gives rise to a stochastic process: An agent observing a certain state chooses an action to cause a state transition with the corresponding reward. By subsequently observing state transitions with corresponding rewards the agent aims to learn an optimal behavior that earns the maximal possible cumulative reward over time. Formally, with the stochastic variables $\left(y(x,a), r(x,a)\right)$ distributed according to $P_0(\cdot{}|x,a)$ the expected immediate reward for each choice of action is given by $\mathbb{E}[r(x,a)]$.
In the following, for a stochastic variable $v$ the notation $v\sim D$ indicates that $v$ is distributed according to $D$. During the stochastic process $(x_{t+1}, r_{t+1})\sim P(\cdot | x_t, a_t)$ the aim of an agent is to maximize the total discounted sum of rewards
\begin{align}
	\mathcal{R} = \sum_{t=0}^{\infty}\gamma^t r_{t+1},
\end{align}
where $\gamma \in (0,1)$ indicates a discount factor that prioritizes rewards in the near future. The choice of action $a_t$ an agent makes in reaction to observing state $x_t$ is determined by its policy $a_t \sim \pi(\cdot | x_t)$. The policy $\pi$ maps observed states to actions and therefore determines the behavior of the agent. Let
\begin{align}
	Q^{\pi}(x,a) = \mathbb{E}\left[ \sum_{t=0}^{\infty} \gamma^t r_{t+1}| x_0 =x, a_0 = a \right]
\end{align}
denote the expected cumulative reward for an agent that behaves according to policy $\pi$. Then we can reduce our problem of approximating the best policy to approximating the optimal $Q$ function. One practical way to achieve this is adjusting $Q$ after each received reward according to
\begin{align}
\label{eqn:q_update}
	Q(x_t, a_t) \leftarrow \ &Q(x_t, a_t) \\ &+ \alpha \left( r_t + \gamma \max_a Q(x_{t+1},a) - Q(x_{t},a_t)    \right),
\end{align}
where $\alpha \in (0,1]$ indicates the learning rate. The process in this setting works as follows: The agent observes a state $x_t$, performs the action $a_t = \arg \max_a Q(x_t, a)$
(where $\arg \max_a f(a)$ denotes the argument value $a$ that maximizes $f(a)$) that maximizes the total expected future reward and thereby causes a state transition from $x_t$ to $x_{t+1}$. Receiving reward $r_t$ and observing $x_{t+1}$ the agent then considers the best possible action $a_{t+1} = \arg \max_a Q(x_{t+1}, a)$. Based on this consideration, the agent updates the value $Q(x_t, a_t)$. If for example the decision of taking action $a_t$ in state $x_t$ led to a state $x_{t+1}$ that allows to choose a high reward action and additionally invoked a high reward $r_t$, the $Q$ value for this decision is adapted accordingly. Here, the factor $\alpha$ determines the rate of this $Q$ function update.

For small state and action spaces, $Q$ can be represented as a table. However, for large state spaces we have to approximate $Q$ with an appropriate function. An approximation using deep neural networks was recently introduced by Mnih et al. \cite{mnih2015human}. For such a representation, the update rule in Equation (\ref{eqn:q_update}) directly translates to minimizing the loss function
\begin{align}
\label{eqn:rf_qloss}
	L = \left( r+ \gamma \max_a Q(x_{t+1},a) - Q(x_t,a_t) \right)^2.
\end{align}
The learning rate $\alpha$ in Equation (\ref{eqn:q_update}) then corresponds to the rate of stochastic gradient descent during backpropagation.

Deep $Q$-networks have been shown to handle large state spaces efficiently. This allows us to define an end-to-end algorithm directly on raw program inputs, as we will see in the next section.

\section{Modeling Fuzzing as a Markov decision process}
\label{sec:rf_model_definition}
In this section we formalize fuzzing as a reinforcement learning problem using a Markov decision process
by defining states, actions, and rewards in the fuzzing context.

\subsection{States}
We consider the system that the agent learns to interact with to be a given ``seed'' program input. Further, we define the states that the agent observes to be substrings of consecutive symbols within such an input. Formally, let $\Sigma$ denote a finite set of symbols. The set of possible program inputs $\mathcal{I}$ written in this alphabet is then defined by the Kleene closure $\mathcal{I}:=\Sigma^*$. For an input string $x=(x_1,...,x_n) \in \mathcal{I}$ let
\begin{align}
\label{eqn:rf_substrings}
S(x):=\left\lbrace (x_{1+i},...,x_{m+i})\ |\ i\geq0,\ m+i \leq n)   \right\rbrace 
\end{align} 
denote the set of all substrings of $x$. Clearly, $\cup_{x\in \mathcal{I}} S(x) = \mathcal{I}$ holds. We define the states of our Markov decision process to be $\mathcal{I}$. In the following, $x\in \mathcal{I}$ denotes an input for the target program and $x'\in S(x)\subset \mathcal{I}$ a substring of this input.

\subsection{Actions}
We define the set of possible actions $\mathcal{A}$ of our Markov decision process to be random variables mapping substrings of an input to probabilistic rewriting rules
\begin{align}
\label{eqn:rf_action_space}
	\mathcal{A} := \left\lbrace  a:\mathcal{I} \rightarrow (\mathcal{I} \times \mathcal{I},\ \mathcal{F}, P) \ |\ a \sim \pi(x') \right\rbrace,
\end{align}
where $\mathcal{F}=\sigma(\mathcal{I} \times \mathcal{I})$ denotes the $\sigma$-algebra of the sample space $(\mathcal{I} \times \mathcal{I})$ and $P$ gives the probability for a given rewrite rule. In our implementation (see Section \ref{sec:rf_implementation}) we define a small subset $A \subset \mathcal{A}$ of probabilistic string rewrite rules that operate on a given seed input.

\subsection{Rewards}
\label{sec:rf_rewards}
We define rewards {\em independently} for both characteristics of: 1) the next performed action $a$ and 2) the program execution with the next generated input $x$, i.e., $r(x,a) = E(x) + G(a)$.


In our implementation in Section \ref{sec:rf_implementation} we experiment with $E$ providing number of newly discovered basic blocks, execution path length, and execution time of the target that processes the input $x$.
For example, we can define the number of newly discovered blocks as
\begin{align}
\label{eqn:rf_bbl_new}
	E_1(x, I') := \left| B(c_x) \setminus \left(  \bigcup_{\chi \in \mathcal{I'}} B(c_{\chi})   \right)   \right|.
\end{align}
where $c_x$ denotes the execution path the target program takes when processing input $x$, $B(c_x)$ is the set of unique basic blocks of this path, and $I'\subset I$ is the set of previously processed inputs. Here, we define a basic block as a sequence of program instructions without branch instructions.


\section{Reinforcement Fuzzing Algorithm}
\label{sec:rf_fuzzing_algorithm}

In this section we present the overall reinforcement fuzzing algorithm.

\subsection{Initialization}

We start with an initial seed input $x \in \mathcal{I}$. The choice of $x$ is not constrained in any way, it may not even be valid with regard to the input format of the target program. Next, we initialize the $Q$ function. For this, we apply a deep neural net that maps states to the estimated $Q$ values of each action, i.e., we simultaneously approximate the $Q$ values for all actions $A$ given a state $x'\in S(x)$ as defined in Equation (\ref{eqn:rf_substrings}). The $x' \mapsto Q(x',a)$ representation provides the advantage that we only need one forward pass to simultaneously receive the $Q$ values for all actions $a\in A$ instead of $|A|$ forward passes. During $Q$ function initialization we distribute the network weights randomly.

\subsection{State Extraction}

The state extraction step \textit{State()} takes as input a seed $x\in \mathcal{I}$ and outputs a substring of $x'\in S(x)$. In Section \ref{sec:rf_model_definition} we defined the states of our Markov decision process to be $\mathcal{I}=\Sigma^*$. For the given seed $x\in \mathcal{I}$ we extract a strict substring $x'\in S(x)$ at offset $o \in \left\lbrace 0,...,|x|-|x'| \right\rbrace $ of width $|x'|$. In other words, the seed $x$ corresponds to the system as depicted in Figure \ref{fig:rf_architecture} and the reinforcement agent observes a fragment of the whole system via the substring $x'$. We experimented with controllable (via action) and predefined choices of offsets and substring widths, as discussed in Section \ref{sec:rf_implementation}.

\subsection{Action Selection}
The action selection step takes as input the current $Q$ function and an observed state $x'$ and outputs an action $a\in A$ as defined in Equation (\ref{eqn:rf_action_space}). Actions are selected according to the policy $\pi$ following an $\epsilon$-greedy behavior: With probability $1-\epsilon$ (for a small $\epsilon>0$) the agent selects an action $a = \arg \max_{a'} Q(x',a')$ that is currently estimated optimal by the $Q$-function, i.e., it exploits the best possible choice based on experience. With a probability $\epsilon$ it explores any other action, where the probability of choice is uniformly distributed within $|A|$.

\subsection{Mutation}
The mutation step takes as input a seed $x$ and an action $a$. It outputs the string that is generated by applying action $a$ on $x$. As indicated in Equation (\ref{eqn:rf_action_space}) we define actions to be mappings to probabilistic rewriting rules and not rewriting rules on their own. So applying action $a$ on $x$ means that we mutate $x$ according to the rewrite rule mapped by $a$ within the probability space $(\mathcal{I} \times \mathcal{I},\ \mathcal{F}, P)$. We make this separation to distinguish between the random nature of choice for the action $a \sim \pi(\cdot | x')$ and the randomness within the rewrite rule.


\subsection{Reward Evaluation}
The reward evaluation step takes as input the target program $P$, an action $a \in A$, and an input $x\in \mathcal{I}$ that was generated by the application of $a$ on a seed. It outputs a positive number $r \in \mathbb{R^+}$. The stochastic reward variable $r(x,a) = E(x) + G(a)$ sums up the rewards for both generated input and selected action. Function $E$ rewards characteristics recorded during the program execution as defined in Section~\ref{sec:rf_rewards}.

\subsection{$Q$-Update}
The $Q$-update step takes as input the extracted substring $x'\in S(x)$, the action $a$ that generated $x$, the evaluated reward $r \in \mathbb{R^+}$, and the $Q$ function approximation, which in our case is a deep neural network. It outputs the updated $Q$ approximation. As indicated above, the choice of applying a deep neural network $Q$ is motivated by the requirement to learn on raw substrings $x'\in S(x)$. The $Q$ function predicts for a given state the expected rewards for all defined actions of $A$ simultaneously, i.e., it maps substrings according to $x' \mapsto Q(x',a)$. We update $Q$ in the sense that we adapt the predicted reward value $Q(x_t,a_t)$ according to the target $r+ \gamma \max_a Q(x_{t+1},a)$. This yields the loss function $L$ given by Equation (\ref{eqn:rf_qloss}) for action $a_t$. All other actions $A\setminus\left\lbrace a_t \right\rbrace$ are updated with zero loss. The convergence rate of $Q$ is primarily determined by the learning rate of stochastic gradient descent during backpropagation as well as the choice of $\gamma$.

\subsection{Joining the Pieces}
\begin{figure}
	\centering
	\includegraphics[scale=.48]{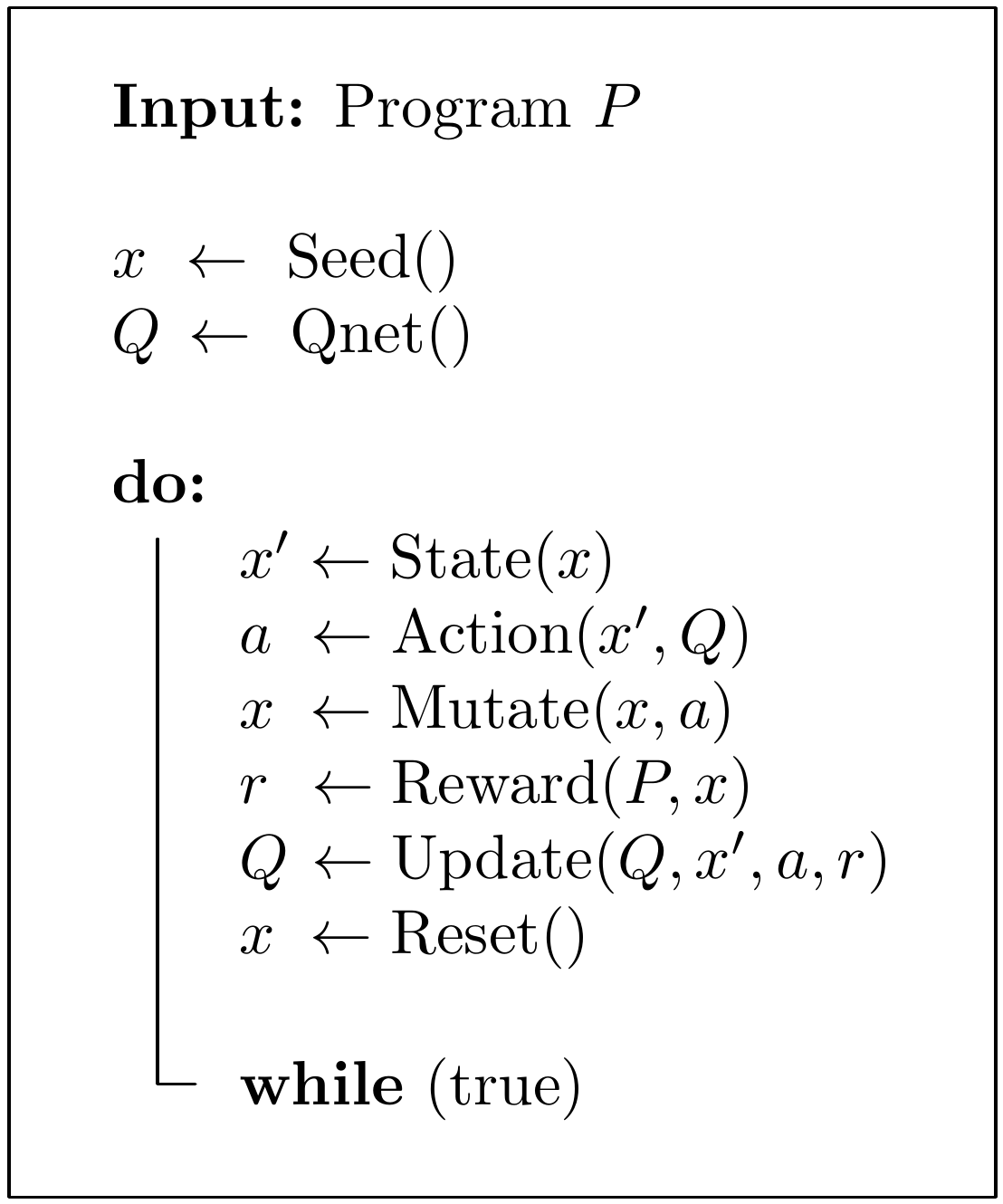}
	\caption{Reinforcement fuzzing algorithm.}
	\label{fig:rf_algorithm}
\end{figure}

Now that we have presented all individual steps we can proceed with combining them to get the overall fuzzing algorithm as depicted in Figure \ref{fig:rf_algorithm}.

We start with an initialization phase that outputs a seed $x$ as well as the initial version of $Q$. Then, the fuzzer enters the loop of state extraction, action selection, input mutation, reward evaluation, $Q$ update, and test case reset. Starting with a seed $x\in \mathcal{I}$, the algorithm extracts a substring $x'\in S(x)$ and based on the observed state $x'$ then chooses the next action according to its policy. The choice is made looking at the best possible reward predicted via $x' \mapsto Q(x',a)$ and applying an $\epsilon$-greedy exploitation-exploration strategy. To guarantee initial exploration we initially define a relatively high value for $\epsilon$ and monotonically decrease $\epsilon$ over time until it reaches a final small threshold, from then on it remains constant. The selected action provides a string substitution as indicated in Equation (\ref{eqn:rf_action_space}) which is applied to $x$ for mutation. The generated mutant input is fed into the target program $P$ to evaluate the reward $r$. Together with $Q$, $x$, and $a$, this reward is taken into account for $Q$ update. Finally, the \textit{Reset()} function periodically resets input $x$ to a valid seed. 
In our implementation we reset the seed after each mutation as described in Section~\ref{sec:rf_implementation}. After reset, the algorithm continues the loop.

We formulated the algorithm with just one single input seed. However, we could generalize this to a set of seed inputs by choosing another seed within this set for each iteration of the main loop. 

The algorithm above performs reinforcement fuzzing with activated policy learning. We show in our evaluation in Section~\ref{sec:rf_implementation} that the $Q$-network generalizes on states. This allows us to switch to high-throughput mutant generation with a fixed policy after a sufficiently long training phase.

\section{Implementation and Evaluation}
\label{sec:rf_implementation}

In this section we present details regarding our implementation together with an evaluation of the prototype.

\subsection{Target Programs}
As fuzzing targets we chose programs processing files in the Portable Document Format PDF. This format is complex enough to provide a realistic testbed for evaluation. From the 1,300 pages long PDF specification~\cite{pdf-manual}, we just need the following basic understanding: each PDF document is a sequence of PDF bodies each of which includes three sections named objects, cross-reference table, and trailer.
While our algorithm is defined to be independent of the targeted input format, we used this structure to define fuzzing actions specifically crafted for PDF objects.

Initially we tested different PDF processing programs including the PDF parser in the Microsoft Edge browser on Windows and several command line converters on Linux. All results in the following presentation refer to fuzzing the \textit{pdftotext} program mutating a $168$ kByte seed file with $101$ PDF objects including binary fields.

\subsection{Implementation}
In the following we present details regarding our implementation of the proposed reinforcement fuzzing algorithm. We apply existing frameworks for binary instrumentation and neural network training and implement the core framework including the $Q$-learning module in Python 3.5.

\paragraph{State Implementation}
Our fuzzer observes and mutates input files represented as binary strings. With $\Sigma = \left\lbrace 0,1 \right\rbrace$ we can choose between state representations of different granularity, for example bit or byte representations. We encode the state of a substring $x'$ as the sequence of bytes of this string. Each byte is converted to its corresponding float value when processed by the $Q$ network. As introduced in Section \ref{sec:rf_fuzzing_algorithm} we denote $o \in \left\lbrace 0,...,|x|-|x'| \right\rbrace $ to be the offset of $x'$ and  $w=|x'|$ to be the width of the current state.

\paragraph{Action Implementation}
We implement each action as a function in a Python dictionary. As string rewriting rules we take both probabilistic and deterministic actions into account. In the following we list the action classes we experiment with.

\begin{itemize}
	\item \textit{Random Bit Flips}. This type of action mutates the substring $x'$ with predefined and dynamically adjustable mutation ratios.
	
	\item \textit{Insert Dictionary Tokens}. This action inserts tokens from a predefined dictionary. The tokens in the dictionary consist of ASCII strings extracted from a set of selected seed files.
	\item \textit{Shift Offset and Width}. This type of action shifts the offset and width of the observed substring. Left and right shift take place at the PDF object level. Increasing and decreasing the width take place with byte granularity.
	\item \textit{Shuffle}. We define two actions for shuffling substrings. The first action shuffles bytes within $x'$, the second action shuffles three segments of the PDF object that is located around offset $o$.
	\item \textit{Copy Window}. We define two actions that copy $x'$ to a random offset within $x$. The first action inserts the bytes of $x'$, the second overwrites bytes.
	\item \textit{Delete Window}. This action deletes the observed substring $x'$.
\end{itemize}

\paragraph{Reward Implementation}
For evaluation of the reward $R(x,a)$ we experimented with both coverage and execution time information.


To measure $E(x) = E_1(x, \mathcal{I'})$ as defined in Equation (\ref{eqn:rf_bbl_new}), we used existing instrumentation frameworks. We initially used the Microsoft Nirvana toolset for measuring code coverage for the PDF parser included in Edge. However, to speed up training of the $Q$ net we switched to smaller parser targets. On Linux we implemented a custom Intel PIN-tool plug-in that counts the number of unique basic blocks within the \textit{pdftotext} program.

\paragraph{$Q$ Network Implementation}
We implemented the $Q$ learning module in Tensorflow~\cite{tensorflow} by constructing a feed forward neural network with four layers connected with nonlinear activation functions. The two hidden layers included between 64 and 180 hidden units (depending on the state size) and we applied $tanh$ as activation function. We initialize the weights randomly and uniformly distributed within $w_i \in [0,\ 0.1]$. The initial learning rate of the gradient descent optimizer is set to $0.02$.

\subsection{Evaluation}
In this section we evaluate our implemented prototype. We present improvements to a predefined baseline and also discuss current limitations. All measurements were performed on a Xeon E5-2690 $2.6$ Ghz with $112$ GB of RAM. The summary of the improvements obtained in accumulated rewards based on different reward functions, modifying state size, and generalization to new inputs is shown in Table~\ref{resultstable}. We now explain the results in more detail.

\subsubsection{Baseline}
\label{sec:baseline}
To show that our new reinforcement learning algorithm actually learns to perform high-reward actions given an input observation, we define a comparison baseline policy that randomly selects actions, where the choice is uniformly distributed among the action space $A$.
Formally, actions in the baseline policy $\pi_B$ are distributed uniformly according to $a \sim \pi_B(\cdot | x)$ and $\forall a\in A:\ \pi_B(a | x)=|A|^{-1}.$ After $n_g=1000$ generations, we calculated the quotient of the most recent $500$ accumulated rewards by our algorithm and the baseline to measure the relative improvement.

\subsubsection{Replay Memory}
We experimented with two types of agent memory: The recorded state-action-reward-state sequences as well as the history of previously discovered basic blocks.
The first type of memory is established during the fuzzing process by storing sequences $e_t:=(x_t,a_t,r_t,x_{t+1})$ in order to regularly replay samples of them in the $Q$-update step. For each replay step at time $t$ a random experience out of $\left\lbrace e_1,...,e_t \right\rbrace$ is sampled to train the $Q$ network.
We could not measure any improvement compared to the baseline with this method.
Second, comparing against the history of previously discovered basic blocks also did not result in any improvement.
Only a memoryless choice of $I'=\emptyset$ yielded good results.
Regarding our algorithm as depicted in Figure \ref{fig:rf_algorithm} we reset the basic block history after each step via the $\textit{Reset}()$ function.

\begin{table}
\centering
\begin{tabular}{|l|r|}
\hline
 & \textbf{Improvement} \\
\hline
\multicolumn{2}{|l|}{Reward functions}\\ \hline
Code coverage $r_1$ & 7.75\% \\ \hline
Execution time $r_2$ & 7\% \\ \hline 
Combined $r_3$ & 11.3\%\\ \hline
\multicolumn{2}{|l|}{State width $w=|x'|$}\\ \hline
$r_2$ with $w = 32$ Bytes & 7\%\\ \hline
$r_2$ with $w = 80$ Bytes & 3.1\%\\ \hline
\multicolumn{2}{|l|}{Generalization to new inputs} \\ \hline
$r_2$ for new input $x$ & 4.7\% \\ \hline
\end{tabular}
\caption{The improvements compared to the baseline (as defined in\ref{sec:baseline}) in the most recent 500 accumulated rewards after training the models for 1000 generations.}
\label{resultstable}
\end{table}


Since both types of agent memory did not yield any improvement, we switched them off for the following measurements. Further, we deactivated all actions that do not mutate the seed input, e.g. random bit flip actions of adjusting the global mutation ratio or shifting offsets and state widths. Instead of active offset $o$ and state width $w=|x'|$ selection via an agent action, we set the offset for each iteration randomly, where the choice is uniformly distributed within $\left\lbrace 0,...,|x|-|x'| \right\rbrace $ and fixed $w = 32$ Bytes.  

\subsubsection{Choices of Rewards}
We experimented with three different types of rewards: Maximization of code coverage $r_1(x,a) = E_1(x,\left\lbrace \right\rbrace )$, execution time $r_2(x,a) = E_2(x)=T(x)$, and a combined reward $r_3(x,a)=E_1(x,\left\lbrace \right\rbrace ) +T(x)$ with rescaled time for multi-goal fuzzing. While $r_1(x,a)$ is deterministic, $r_2(x,a)$ comes with minor noise in the time measurement. Measuring the execution time for different seeds and mutations revealed a variance that is two orders of magnitude smaller than the respective mean so that $r_2$ is stable enough to serve as a reliable reward function. All three choices provided improvements with respect to the baseline.

When rewarding execution time according to $r_2$ our proposed fuzzing algorithm cumulates in average $7\%$ higher execution time reward in comparison to the baseline.
%

Since both time and coverage rewards yielded comparable improvements with regard to the baseline, we tested to what extend those two types of rewards correlate: We measured an average Pearson correlation coefficient of $0.48$ between coverage $r_1$ and execution time $r_2$. This correlation motivates the combined reward $r_3(x,a)=E_1(x,\left\lbrace \right\rbrace ) +T(x)$, where $T(x)$ is a simple rescaling of execution time by a multiplicative factor $1*10^{6}$ so that the execution time contributes to the reward equitable to $E_1$. Training the $Q$ net with $r_3$ yielded an improvement of $11.3\%$ in execution time. This result is better that taking exclusively $r_1$ or $r_2$ into account. There are two likely explanations for this result. First, the noise of time measurement could introduce rewarding explorative behavior of the $Q$ net. 
Second, deterministic coverage information could add stability to $r_2$.


\subsubsection{$Q$-net Activation Functions}
From all activation functions provided by the Tensorflow framework, we found the $tanh$ function to yield the best results for our setting. The following list compares the different activation functions with respect to improvement in reward $r_1$.
\begin{center}
	\begin{tabular}{|c|c|c|c|c|c|}
		\hline
		tanh & sigmoid & elu & softplus & softsign &relu \\ \hline
		7.75\% & 6.56\% & 5.3\% & 2\% & 6.4\% & 1.3\% \\ \hline
	\end{tabular}
\end{center}

\subsubsection{State Width}
Increasing the state width $w=|x'|$ from $32$ Bytes to $80$ Bytes decreased the improvement (measured in average reward $r_2(x,a)$ compared to the baseline) from $7\%$ to $3.1\%$. In other words, smaller substrings are better recognized than large ones. This indicates that our proposed algorithm actually takes the structure of the state into account and learns to perform best rewarded actions according to this specific structure.



\subsubsection{State Generalization}

In order to achieve high-throughput fuzzing we tested if the already trained $Q$ net generalizes to previously unseen inputs. This would allow us to switch off $Q$ net training after a while and therefore avoid the high processing costs of evaluating the coverage reward. To measure generalization we restricted the offset $o \in \left\lbrace 0,...,|x|-|x'| \right\rbrace $ in the training phase to values in the first half of the seed file. For testing, we omitted reward measurement in the $Q$ update step as depicted in Figure \ref{fig:rf_algorithm} to stop the training phase and only considered offsets in the second half of the seed file. This way, the $Q$ net is confronted with previously unseen states. This resulted in an improvement in execution time of $4.7\%$ compared to the baseline.

\section{Conclusion}
\label{sec:rf_conclusion}
Inspired by the similar nature of feedback-driven random testing and reinforcement learning, we introduce the first fuzzer that uses reinforcement learning in order to learn high-reward mutations with respect to predefined reward metrics. By automatically rewarding runtime characteristics of the target program to be tested, we obtain new inputs that likely drive program execution towards a predefined goal, such as maximized code coverage or processing time. To achieve this, we formalize fuzzing as a reinforcement learning problem using Markov decision processes. This allows us to construct an reinforcement-learning fuzzing algorithm based on deep $Q$-learning that chooses high-reward actions given an input seed. 

The policy $\pi$ as defined in Section \ref{sec:rf_background} can be viewed as a form of generalized grammar for the input structure. Given a specific state, it suggests a string replacement (i.e., a fuzzing action) based on experience. Especially if we reward execution path depth, we indirectly reward validity of inputs with regard to the input structure, as non-valid inputs are likely to be rejected early during parsing and result in small path depths. We presented preliminary empirical evidence that our reinforcement fuzzing algorithm can learn how to improve its effectiveness at generating new inputs based on successive feedback. Future research should investigate this further, with more setup variants, benchmarks, and experiments.





\bibliographystyle{IEEEtran}
\bibliography{archive}
%
%
%

\end{document}